\documentclass[10pt,twocolumn,letterpaper]{article}

\usepackage{cvpr}
\usepackage{times}
\usepackage{epsfig}
\usepackage{graphicx}
\usepackage{amsmath}
\usepackage{amssymb}
\usepackage{mathtools}
\usepackage{siunitx}
\usepackage[flushleft]{threeparttable}
\usepackage{float}
\usepackage{caption,subcaption}

\usepackage{caption}
\floatstyle{plaintop}
\restylefloat{table}
\DeclareMathOperator*{\argmin}{argmin}

\usepackage[pagebackref=true,breaklinks=true,letterpaper=true,colorlinks,bookmarks=false]{hyperref}
\usepackage{multirow}
\cvprfinalcopy 

\newcommand{\RNum}[1]{\uppercase\expandafter{\romannumeral #1\relax}}
\ifcvprfinal\fi

\begin{document}
	
	\title{Deep 3D Face Identification}
	
	\author{Donghyun Kim\qquad Matthias Hernandez\qquad Jongmoo Choi\qquad G\'erard Medioni\\
		USC Institute for Robotics and Intelligent Systems (IRIS)\\
		Unversity of Southern California \\
		{\tt\small \{kim207, mthernan, jongmooc, medioni\}@usc.edu}
	}
	
	\maketitle
	
	\begin{abstract}
		We propose a novel 3D face recognition algorithm using a deep convolutional neural network (DCNN) and a 3D augmentation technique. The performance of 2D face recognition algorithms has significantly increased by leveraging the representational power of deep neural networks and the use of large-scale labeled training data. As opposed to 2D face recognition, training discriminative deep features for 3D face recognition is very difficult due to the lack of large-scale 3D face datasets. In this paper, we show that transfer learning from a CNN trained on 2D face images can effectively work for 3D face recognition by fine-tuning the CNN with a relatively small number of 3D facial scans. We also propose a 3D face augmentation technique which synthesizes a number of different facial expressions from a single 3D face scan. Our proposed method shows excellent recognition results on Bosphorus, BU-3DFE, and 3D-TEC datasets, without using hand-crafted features. The 3D identification using our deep features also scales well for large databases.  
	\end{abstract}
	
	\section{Introduction}
	
	Face recognition has been an active research topic for many years. It is a challenging problem because the facial appearance and surface of a person can be vary greatly due to changes in pose, illumination, make-up, expression or hard occlusions.
	
	Recently, the performance of 2D face recognition systems~\cite{parkhi2015deep, schroff2015facenet,taigman2014deepface}	was boosted significantly with the popularization of deep convolutional neural networks (CNN). It turns out that recent methods using CNN feature extractors trained on a massive dataset outperform conventional methods using hand-crafted feature extractors, such as Local Binary Pattern~\cite{ahonen2004face} or Fisher vectors~\cite{simonyan2013fisher}. Deep learning approaches require a large dataset to learn a face representation which is invariant to different factors, such as expressions or poses. Large scale datasets of 2D face images can be easily obtained from the web. FaceNet~\cite{schroff2015facenet} uses about 200M face images of 8M independent people as training data. VGG Face~\cite{parkhi2015deep} assembled a massive training dataset containing 2.6M face images over 2.7K identities.
	
	With 3D modalities, recent research \cite{lei2016two, li2015towards, spreeuwers2011fast} has focused on finding robust feature points and descriptors based on geometric information of a 3D face in a hand-crafted manner. Those methods achieve good recognition performances but involve relatively complex algorithmic operations to detect key feature points and descriptors as compared to end-to-end deep learning models. While some of these methods can do verification in real-time, they often do not scale well for identification tasks where a probe scan needs to be matched with a large-scale gallery set. 
	\begin{figure}[t]
		\centering
		\includegraphics[width = 0.5\textwidth]{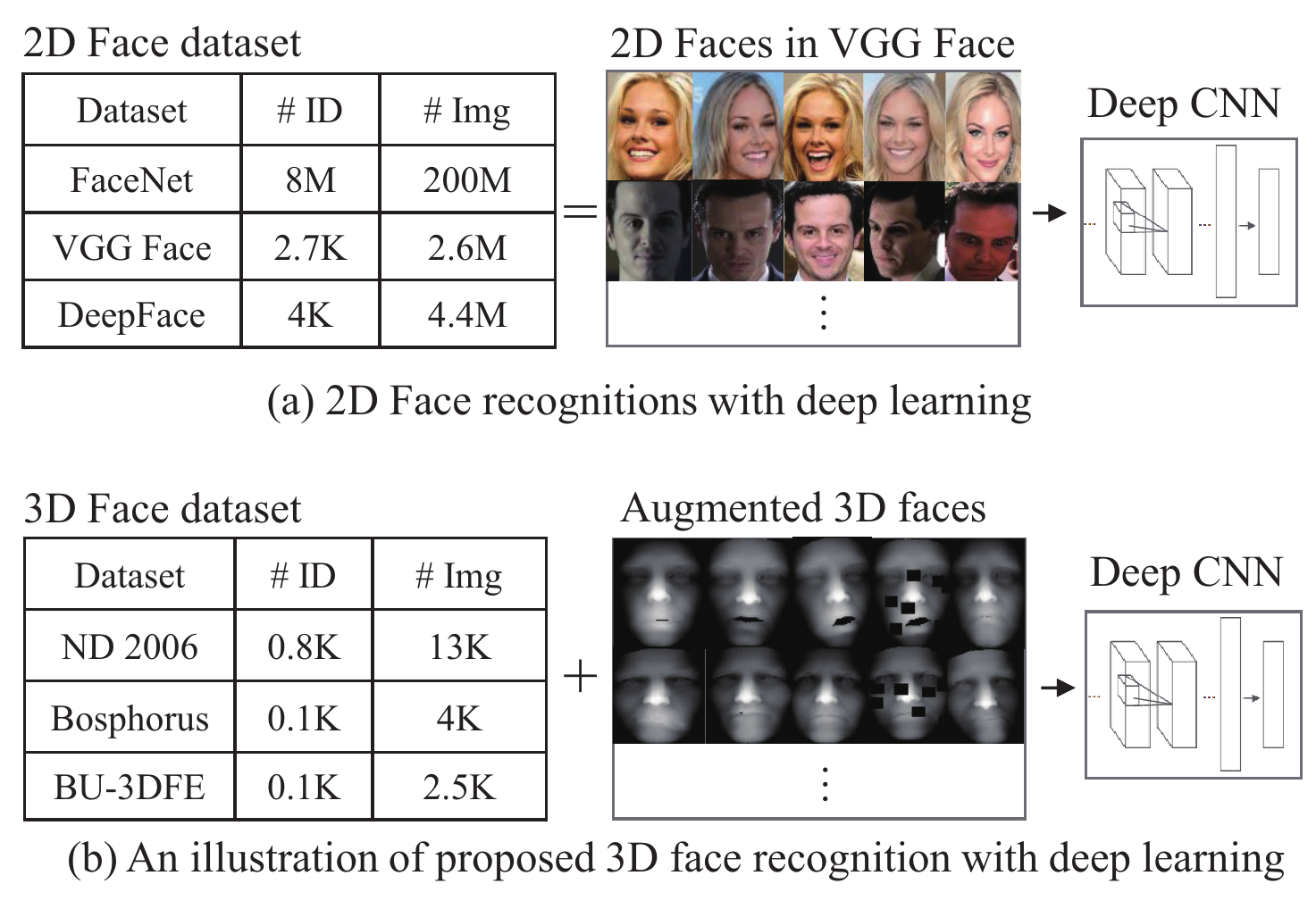}
		\caption{Challenges in 3D face recognition with deep learning due to the absence of massive datasets. (a) Datasets for 2D are large enough (200M at most) to train a DCNN. Images in VGG Face contain rich variations of expression, pose, occlusion, and illumination. (b) The number of 3D images is very limited (13K at most). 
			Therefore, it is important to augment 3D faces to add variations for leaning a robust representation. }
		\label{frsystem}
	\end{figure} 
	Compared to publicly available 2D face databases, 3D scans are hard to acquire,  and the number of scans and subjects in public 3D face databases is limited. According to the survey in~\cite{patil20153}, the biggest 3D dataset is ND 2006~\cite{faltemier2007using} which contains 13,450 scans over 888 individuals. It is small compared to publicly available labeled 2D faces, and may not be sufficient to train a deep convolutional neural network from scratch. Figure \ref{frsystem} shows available face datasets for both 2D and 3D and exhibits the challenges of 3D face recognition with deep learning. 
	As a result, coping with the limited amount of available 3D data is challenging. 
	We propose to leverage existing networks, trained for 2D face recognition, and fine-tune them with a small amount of 3D scans in order to perform 3D to 3D surface matching. 
	
	
	Another challenge intrinsic to the recognition tasks comes from the need to minimize intra-class variances (e.g., differences in the same individual under expression variations) while maximizing inter-class variances (differences between persons). For faces, variations in expression impact the 3D structure and can degrade the performance of recognition systems~\cite{patil20153}. 
	To address this issue, we propose to augment our 3D face database with synthesized 3D face data considering facial expressions. To augment training data, we use multi-linear 3D morphable models in which the shape comes from the Basel Face Model~\cite{paysan20093d}, and the expression comes from FaceWarehouse~\cite{cao2014}. 
	
\begin{figure*}[t]
		\centering
		\includegraphics[width = 0.85\textwidth]{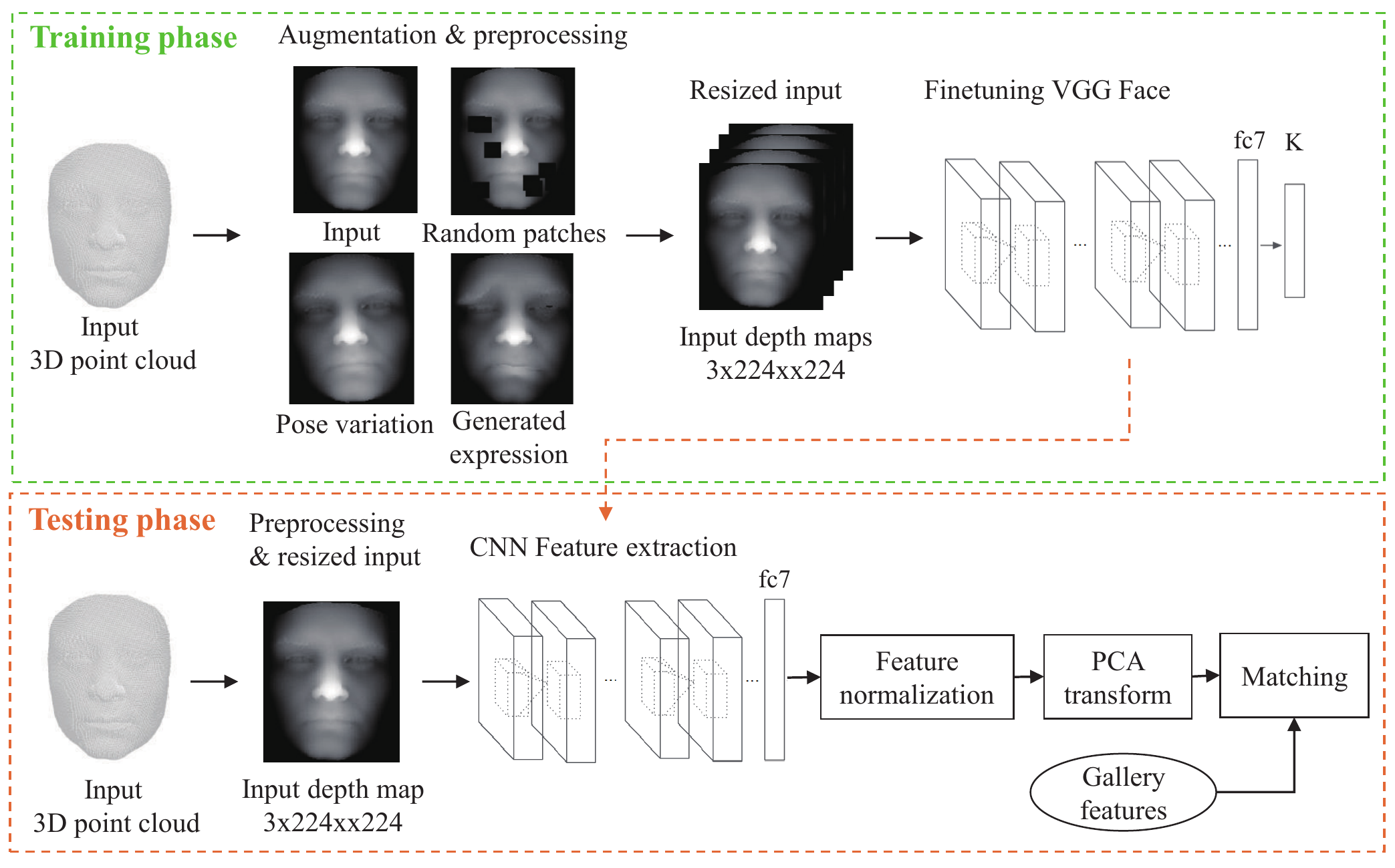}
		\caption{An overview of the proposed face identification system. In the training phase, we align 3D facial point clouds with a reference model, augment the point clouds, and convert them to 2D depth maps. Depth maps are resized to fit the size of VGG Face. In the testing phase, a probe scan is preprocessed and resized. Then, a face representation is extracted from the fine-tuned CNN. After normalization of features and Principal Component Analysis transform, one's identity is determined by the matching step.}
		\label{overview}
	\end{figure*}

	To pass our 3D data to the 2D-trained CNN, we project the point clouds onto a 2D image plane with an orthographic projection. To make our system robust to small alignment error, each 3D shape is augmented by rigid transformations: 3D rotations and translations before the projection. Some random patches are also added to the 3D data to simulate random occlusions (e.g., facial hair, covering by hands or artifacts). We fine-tune a deep CNN trained for 2D face recognition, VGG-Face~\cite{parkhi2015deep}, with the augmented data. Figure \ref{overview} illustrates our proposed method. We report performances on standard public 3D databases: Bosphorus~\cite{savran2008bosphorus}, BU-3DFE~\cite{yin20063d}, and 3D-TEC~\cite{vijayan2011twins}.
	
	Our contributions are as follows:
	\begin{enumerate}
		\item  To our knowledge, this work is the first to use a deep convolutional neural network for 3D face recognition. We frontalize a 3D scan, generate a 2.5D depth map, extract deep features to represent the 3D surface, and match the feature vector to perform 3D face recognition. 
		\item  We propose a 3D face augmentation method that generates a number of person specific 3D shapes with expression changes from a single raw 3D scan, which allows us to enlarge a limited 3D dataset and improve the performance of 3D face recognition in the presence of expression variations.
		\item  We have validated our approach on 3 standard datasets. Our method shows comparable results to the state of the art algorithms while enabling efficient 3D matching for large-scale galleries.
	\end{enumerate}
	
	An overview of our framework is presented in Figure~\ref{overview}. The rest of the paper is organized as follows: Section \ref{RelatedWork} reviews the related work. Section \ref{Method} describes our proposed method. Our augmentation and performances on the public 3D databases are evaluated in Section \ref{Experiments}. Section \ref{Conclusions} concludes the paper.
	
	\section{Related work} \label{RelatedWork}
	We review prior works on 3D face recognition, 2D face recognition using deep convolutional neural networks (DCNN), and the use of CNNs for 3D object recognition.
	
	{\bf3D face recognition} An overview of 3D face recognition systems is presented in~\cite{patil20153}. Challenges in 3D face recognition come from variations in expression, occlusions, and missing parts. Most recent methods \cite{lei2016two, li2014expression,li2015towards}
	extract hand-crafted features from a 3D face and match the features with other features using a metric for verification/identification tasks. Pre-processing steps such as 3D face detection, landmark extraction, and registration may be required depending on the method. Existing methods can be largely classified into two categories: holistic approaches and local-region based approaches. 
	
	In holistic approaches, the whole 3D face is used for recognition. Drira \etal~\cite{drira20133d} proposed a method that uses curvature analysis of the 3D facial surfaces. The method extracts radial curves of facial surfaces around the nose tip and matches radial curves for recognition. To handle missing or occluded parts, they remove occlusions and attempt to restore missing parts. Morphable model-based approaches were introduced in \cite{ter2010expression, ocegueda2011ur3d}. 
	After fitting a morphable model into a probe scan, the fitted 3D face is passed to the recognition module. As each probe scan need to be fitted, these approaches usually are time-consuming.
	
	In local-region based approaches, features are extracted from several local sub-regions and local information. Li \etal~\cite{li2015towards} detect 3D keypoints where local curvatures are high. They propose three keypoint descriptors using the local shape information of detected keypoints. A multi-task sparse representation based on Sparse Representation based Classifier (SRC) is presented in \cite{wright2009robust}.
	Lei \etal~\cite{li2015towards} propose a robust local facial descriptor. After detecting keypoints based on Hotelling transform, they set the descriptor by calculating four types of geometric features in a keypoint area. Then, they use a two-phase classification framework with extracted local descriptors for recognition. 
	
	Due to the complexity of these algorithms, they suffer from slow feature extraction and/or matching processes, which limits their scalability as 
	compared to end-to-end learning systems.
	
	{\bf2D face recognition} 
	Most of the recent work on 2D face recognition~\cite{parkhi2015deep, schroff2015facenet,taigman2014deepface} relies on deep learning approaches using massive datasets. In these approaches, deep neural networks learn 2D face representation directly from 2D images of faces. By training a deep neural network on a large dataset, a robust face representation can be learned without hand-crafted features. Each face image is pre-processed (e.g. face detection, cropping, and alignment) depending on its deep CNN structure. DeepFace~\cite{taigman2014deepface} uses a large photo collection of 4M faces over 4K people from Facebook for training a deep CNN. They align 2D faces using a general 3D shape model and use a siamese network which minimizes the distance between a pair of faces from the same identity and maximizes the distances between a pair of faces from different identities. FaceNet~\cite{schroff2015facenet} also uses a large dataset of 200M faces over 8M identities. Their deep CNN architecture is based on inception modules~\cite{szegedy2015going} and use triplet loss which minimizes the distances between an anchor image and another image from identities but maximizes the distance between the anchor image and the other image from different identities within a set of three images. VGG Face~\cite{parkhi2015deep} proposes a procedure to assemble a large dataset with a small cost and trained VGG-16 nets on the dataset of 2.6M images. This dataset is relatively small compared with \cite{taigman2014deepface, schroff2015facenet}. Masi \etal~\cite{masi2016we} augment face images using 3D rendered images on poses, shapes, and closed mouth expressions using a 3D generic face. A key limitation of these approaches is that training of DCNNs requires a massive and carefully-designed dataset.
	
	{\bf Convolutional neural nets for 3D objects}  
	3D object recognition using deep convolutional neural networks is not well developed. First, 3D objection recognition still suffers from a lack of large annotated 3D database. For example, the most commonly used 3D dataset, ModelNet~\cite{wu20143d}, contains only 150K shapes. To put this in perspective, ImageNet dataset~\cite{deng2009imagenet}, which is a very large 2D image database, contains tens of millions of annotated data. Second, effective representations to pass 3D objects to a CNN are yet to be determined. Representations for 3D objects can be classified into two categories: model-based methods and view-based methods. 
	
	In model-based methods, a whole 3D model is used for feature extraction. Wu \etal~\cite{wu20143d} use a volumetric representation for 3D shapes. Each voxel in a grid contains a binary value depending on a mesh surface and the size of the grid is $30\times30\times30$. They use ModelNet dataset as their training data and Convolutional Deep Belief Network. A $30\times30\times30$ resolution may work for object classification but is not fine enough to represent facial shapes. Using a high enough voxel resolution to capture fine facial structure variations would require massive amounts of memory.
	
	In view-based methods, a set of views from a 3D shape is used for recognition. The advantage of this method is that it can leverage existing 2D image datasets, such as ImageNet. Su \etal~\cite{su2015multi} create multiple 2D image renderings of a 3D object with different camera positions and angles for training and testing data. They use a CNN pre-trained on ImageNet. A first CNN is used for extracting features from each image and combine the features with element-wise maximum operation. A second CNN is used to compact the shape descriptor. 
	
	Since there is no prior research on 3D face recognition using a deep convolutional neural net, there is no known representation for 3D facial scans. Indeed, a $30\times30\times30$ volumetric grid~\cite{wu20143d} may be too coarse to represent a 3D face and multiple views of a face may not be needed.

	\section{Method} \label{Method}
	
	Figure~\ref{overview} shows the proposed face identification process. We represent a 3D facial point cloud with an orthogonally projected 2D depth map. We use 2D depth maps to fine-tune VGG Face, which is pre-trained for the task of face recognition from 2D face images. In the training phase, we augment our 3D data to enlarge the size of our dataset and make the CNN robust. A 3D point cloud of a facial scan is augmented with expression and pose variations. After converting a 3D point cloud into a 2D depth map, random patches are removed from the depth maps and used as augmentation to simulate hard occlusions. 
	In the testing phase, the fine-tuned CNN is used as a feature extractor. We take extracted features from the CNN as a 3D face representation. Then, we determine one's identity by matching its representation with a gallery set.    
	
	
	\subsection{Preprocessing} 
	
	3D scans can be provided in different conditions that can impact the appearance of the rendered images, notably with different poses or with large 3D noise coming from the sensor. We wish to minimize these factors when converting the 3D models to 2D maps.
	
	First, we align all the facial 3D models together using a classical rigid-ICP~\cite{castellani20123d} between each 3D scan and a reference facial model. To initialize the algorithm, we find the nose tip in the 3D point cloud, and crop the point cloud within an empirically set radius ($100 mm$). This process keeps only the facial region and enables better convergence of the ICP algorithm. This process is similar to~\cite{li2014expression}.
	%
	
	The aligned 3D scan is then projected orthographically onto a 2D image to generate a depth map. 
	3D Points are scaled by $200/r$ to create a $200\times200$ size depth 2D map, where $r$ is a constant radius value used for face cropping. For a given 3D point $[x,y,z]^{T}$, the coordinates $(x, y)$ in a 2D depth map may not be integers. 
	We calculate each pixel values using a 
	mean with bilinear interpolation, as described in \cite{hernandez2015near}.
	A 3D point cloud can contain spikes due to sensor noise. Therefore, we apply a classical median filtering to the depth images to generate the final results. 
	\subsection{Augmentation}
	\label{sec:augment}
	We leverage existing CNNs used for 2D face recognition and fine-tune them for 3D face recognition. With a limited amount of available 3D scans, we do not have a lot of variability in the data, which might yield overfitting. To address this issue, we augment our training data with synthesized 3D models.
	
	Previous works on 3D face recognition have shown that the performance is sensitive to variations in expressions and errors in the alignment~\cite{patil20153}. Therefore, designing a system robust to these factors is desirable. However, variation of expressions in the training data (e.g. FRGC) is very limited (e.g. sad, happy, and neutral). We may want to collect more training data which contains richer expressions but acquiring lots of 3D faces needs formidable efforts. Our expression generation method allows us to get a wide variety of expressions with a given dataset.        
	\begin{figure*}[ht]
		\centering
		\begin{subfigure}[b]{0.5\linewidth}
			\centering\includegraphics[width = 1.0\textwidth]{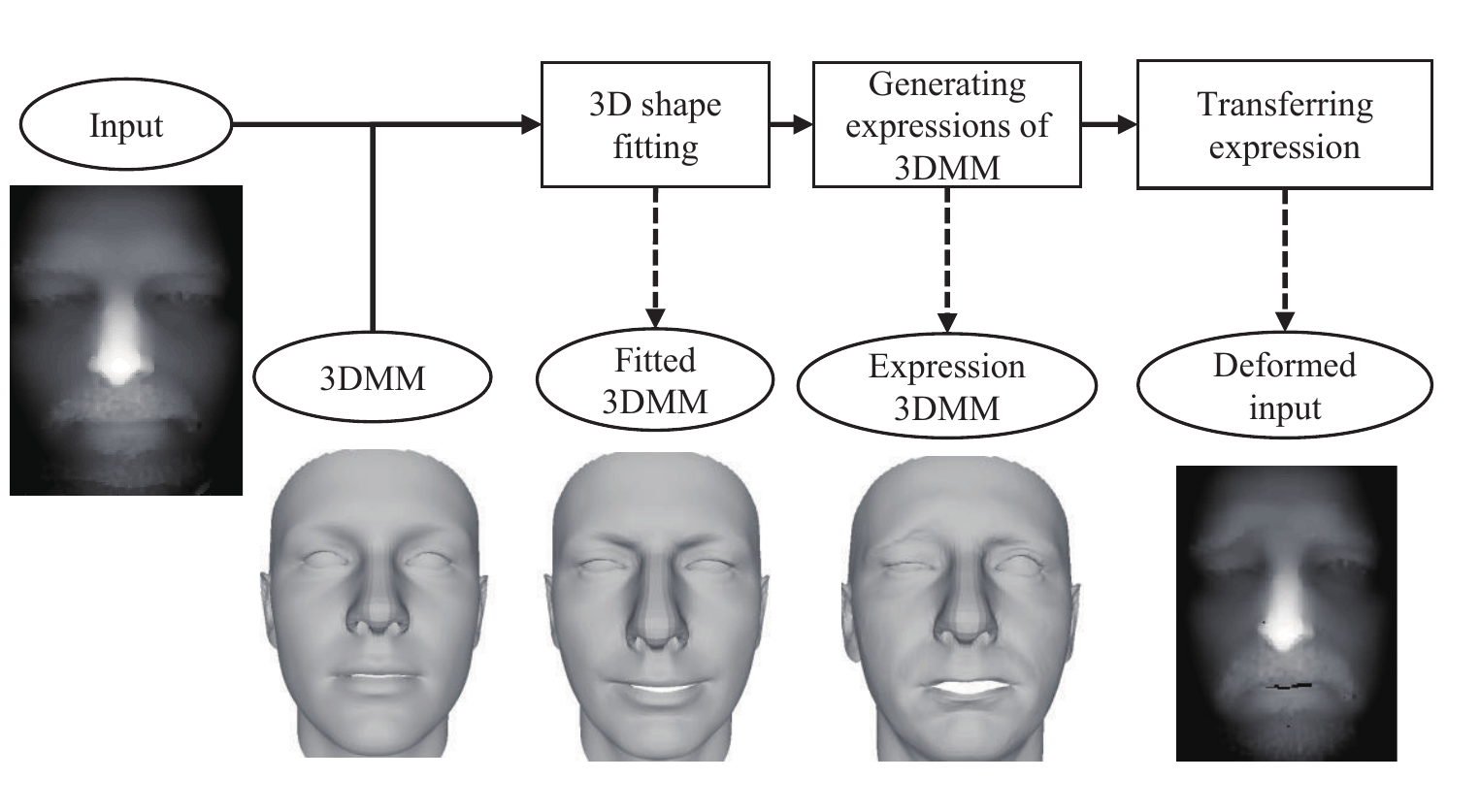}
			\caption{An overview of expression generation method\label{exaug1}}
		\end{subfigure}%
		\begin{subfigure}[b]{0.5\linewidth}
			\centering\includegraphics[width = 1.0\textwidth]{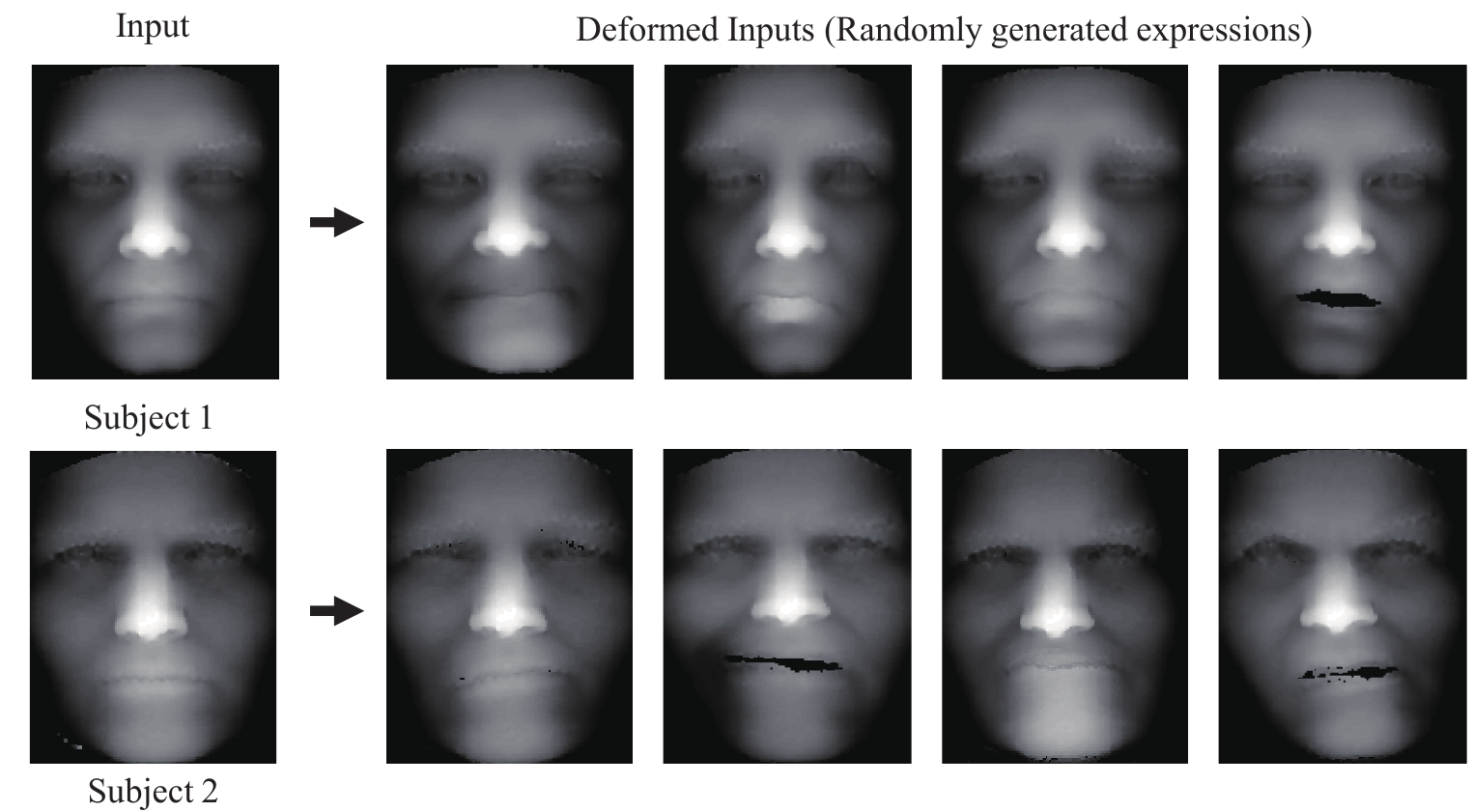}
			\caption{Examples of randomly generated expressions.\label{exaug2}}
		\end{subfigure}
		\caption{An overview of person specific expression generation method and its examples. (a) Input: a facial 3D point cloud, Output: a deformed 3D point cloud (b) The first column represents neutral scans of input faces. Rest columns represent generated individual specific expressions from an input. For visualization, we show a 3D point cloud as a depth map.}
	\end{figure*}
	
	\subsubsection{Expression Generation}
	
	First, we augment our 3D scans with expression variations. To be specific, we modify the expression of every 3D facial scan in the training dataset, and add the resulting point cloud to the training dataset. 
	
	Figure \ref{exaug1} shows an overview of the expression augmentation method. Adding expressions to a facial point scan consists of three steps: (1) 3D shape fitting a 3D morphable model (3DMM) to an input scan (3D point cloud), (2) adding expressions to the 3DMM, and (3) transferring expressions of the 3DMM to the input. Figure \ref{exaug2} shows examples of randomly generated expressions for two subjects.
	
	{\bf 3DMM} We use a multi-linear 3D morphable model (3DMM), which contains variations in both shape and expression. The shape information comes from the Basel Face Model (BFM)~\cite{paysan20093d}, while the expression comes from FaceWarehouse~\cite{cao2014}. The 3DMM represents a face by two vectors $\alpha$ and $\beta$, which represent the shape and the expressions, respectively. A 3D point cloud can then be computed with a linear equation:
	
	\begin{equation}
	\mathbf{X} = \mathbf{\bar{X}} + \mathbf{P}_{s} \alpha + \mathbf{P}_{e} \beta, 
	\label{eq:3DMM}
	\end{equation}
	
	where $ \mathbf{\bar{X}} $ is the average facial point cloud, $\mathbf{P}_{s}$ is provided by the BFM, and $\mathbf{P}_{e}$ by FaceWarehouse.
	
	{\bf 3D shape fitting} To fit a 3DMM to an input scan, we iteratively update shape, pose, and expression in an iterative-closest point manner. The fitting process is similar to \cite{amberg2008expression}. It uses only a 3D point cloud for fitting. In our experiment, we fit the 3DMM to neutral expression scans.
	
	{\bf Generating expressions} We can add random expressions to the fitted 3DMM by randomly varying values of the expression $\beta$ parameters in the 3DMM. To be specific, our 3DMM has $29$ expression parameters. In order to generate various expressions, we generate a set of random vectors $\beta$, which uses a random number of expression parameters and assign random values each time. We limit a parameter value $\beta_{i}$ within a range 
	($-0.05<\beta_{i}<0.05$) to generate natural looking expressions.
	
	{\bf Transferring expressions} The results of 3DMM cannot be used directly as training data. Indeed, 3DMMs are smooth by design and smooth out a lot of local information that may be important for recognition. As a result, we propose to only use the 3D morphable model to compute the deformation field between the original point cloud and an expression-augmented point cloud.
	
	We first define a target expression by a randomly generated 3D model using the 3DMM. We then compute a displacement vector field that maps from the input raw 3D scan to the target augmented 3D data. A displacement vector $\Delta_{i}$ from a 3D point in the fitted 3DMM ($\Omega_{i}$) to the corresponding point of the deformed 3DMM ($\Psi_{i}$) is computed as
	\begin{equation}
	\Delta_{i} = \Psi_{i} - \Omega_{i}
	\end{equation}
	where $\mathbf{\{\Delta_{i}\}}^{N}_{i=1}$ represents a set of displacement vectors from the fitted to the deformed 3DMM and $N$ is the number of points in a 3DMM. To apply the displacement vectors to the input, we take for every input point ($\mathbf{X}_{i}$) a displacement vector based on the nearest point in the fitted 3DMM:
	\begin{equation}
	\begin{split}
	\Delta_{j} &=\argmin_{j} ||X_{i} - \Omega_{j}\||_{2}^2, \\
	X'_{i} &= X_{i} + \Delta_{j},
	\end{split}
	\end{equation}
	where $\mathbf{\Delta}_{j}$ is the displacement vector corresponding to the $\mathbf{X}_{i}$. $\mathbf{\{X'}_{j}\}^{M}_{j=1}$ represents the deformed input where M is the number of point in the input. 
	\subsubsection{Pose Variations}
	Secondly, we augment our data with variations on 3D pose. A well-known problem of the rigid ICP registration is that it does not guarantee an optimal convergence. It means that all 3D faces may not be accurately registered to the reference face and have different poses. Furthermore, a CNN is not invariant to pose transformations \cite{laptev2016ti, jaderberg2015spatial}. Therefore, this augmentation aims at making the CNN invariant to minor pose variations.
	To do so, we simply apply randomly generated rigid transformation matrices ($\mathbf{M} = \mathbf{[R\; t]}$) to an input 3D point cloud. A rotation matrix ($\mathbf{R} \in \mathbb{R}^{3\times3}$) is generated by multiplying yaw, pitch and row rotations ($\mathbf{R} = \mathbf{R_z(\theta_3)R_y(\theta_2)R_x(\theta_1)} $) with random angle degrees ($-10^{\circ} <\theta_1,\theta_2, \theta_3< 10^{\circ}$). A translation vector is also generated with random values ($\mathbf{t} = [x,y,z]^{T} \text{ where} -10<x,y,z< 10$). 
	
	\subsubsection{Random Patches}
	Finally, we put eight $18\times18$ size patches on each 2D depth map at random positions. These random patches are used to prevent overfitting to specific regions of the face. As a result, a patch-augmented CNN learns every region of a face. In 2D face images, we can easily find this kind of training data, such as when the subject is wearing sunglasses or is occluded by other objects. With a 3D face, we simulate occluded data by hiding random patches in the depth map. Note that other types of patches, with different size and shape, are also possible.
	
	\subsection{Fine-tuning}
	
	To build our 3D face recognition system, we start from VGG Face~\cite{parkhi2015deep} that is pre-trained on 2D face images, and fine-tune the network with our augmented 2D depth maps. In order to fit the input size of VGG Face, a depth map is resized to $224\times224$ and 3 channels.
	
	We transfer all the weights from VGG Face but replace the last fully connected layer (FC8) with a new last fully connected layer and a softmax layer. The new last layer has the size of the number of subjects in training data and weights are randomly initialized from a Gaussian distribution with zero mean and standard deviation $0.01$. We use SGD using mini-batches of 32 samples and set a learning rate $0.001$ for pre-trained layers but $0.01$ for the last layer.
	
	\subsection{Identification}
	The fine-tuned CNN is used for extracting features. We take a 4096-dimensional vector from the FC7 layer as a face representation. A feature vector is normalized by taking the square root of each element of the feature vector. After acquiring features of gallery and probe set, we perform Principal Component Analysis on features from a probe set with features from a gallery set. In a matching step, we calculate a cosine distance between features from a probe and a galley. One's identity is determined by the closest gallery feature.
	
	\section{Experiments} \label{Experiments}
	
	\subsection{3D Face Databases} We use the augmented FRGC~\cite{phillips2005overview} (FRGC v1 and FRGC v2) and CASIA 3D~\cite{CAISA3D} as our training data. Each gallery set of the databases except 3D-TEC~\cite{vijayan2011twins} is augmented and used for training data. Performances are evaluated on the Bosphorus~\cite{savran2008bosphorus}, BU-3DFE ~\cite{yin20063d} and 3D-TEC. 
	
	{\bf FRGC v1, v2} This database which is a subset of ND 2006~\cite{faltemier2007using} consists of 4,950 3D facial scans of 577 subjects. Although the number of identities is relatively large compared with other databases, it contains limited expression variations. 
	We use this database as our training data.
	
	{\bf CASIA 3D} The Casia 3D database contains 4,624 scans of 123 subjects. 
	Each subject was captured with different expressions and poses. We only use scans with expressions, which amount to 3,937 scans over 123 subjects. This database is used as a training set.
	
	{\bf Bosphorus} The Bosphorus database contains 4,666 3D facial scans over 105 subjects with rich expression variations, poses, occlusions. 
	The 2,902 scans contain expression variations from 105 subjects. In the experiment, 105 first neutral scans from each identity are used as a gallery set and 2797 non-neutral scans are used as a probe set.
	
	{\bf BU-3DFE} The BU-3DFE database contains 2500 3D facial expression models of 100 subjects. Each subject made 6 expressions (e.g. happiness, disgust, fear and so on) with four levels of intensity from low to high and a neutral expression. The resolution of this database is low compared to other databases. 
	
	{\bf 3D-TEC} The 3D-TEC database contains 107 pairs (total 214 subjects) of twins and expressions of neutral and smiling scan are captured for each subject. Since twins are visually similar to each other, it is harder to identify a probe where the corresponding twin of the probe is in a gallery set. Identifying twins visually are sometimes hard even for humans. The standard protocol for 3D-TEC has four scenarios (Case \RNum{1},\RNum{2},\RNum{3}, and \RNum{4}) described in \cite{vijayan2011twins}. 
	
	\subsection{Analysis of Augmentation}
	As described in section~\ref{sec:augment}, we augment 3D face on expressions, pose variations, and random patches. We evaluate each augmentation method separately. As we only use the FRGC dataset as a training set, the amount of data is so limited that a DCNN may not be trained well from scratch. Therefore, we add shallow convolution layers for analysis of performance. In total, we use three different CNNs to evaluate augmentation methods in detail: (a) shallow net (2 convolution layers and 2 fully connected layer) with random initial weights, (b) VGG-16 with random initial weights, (c) VGG Face which is pre-trained from 2D face images.
	
	{\bf Expression generation}
	When augmenting, we selected the first scan of each identity (577) in the FRGC database and generated 25 expressions with the selected scan. A total of 14,425 (577*25) expression scans were generated.
	
	{\bf Pose variations}
	We apply pose augmentations to every scan in the FRGC dataset. We applied 10 random rigid-body transformations to each scan. The number of augmented data is 49,500 (4950*10) and the total number of training data is 54,450.
	
	{\bf Random patches}
	We augmented a 3D face in the FRGC database by putting random patches on the corresponding 2D depth map. We generated 10 images per a scan. The total number of training data is 54,450 which is the same as in the pose variation experiment.
	\begin{figure}[t]
		\centering
		\includegraphics[width = 0.5\textwidth]{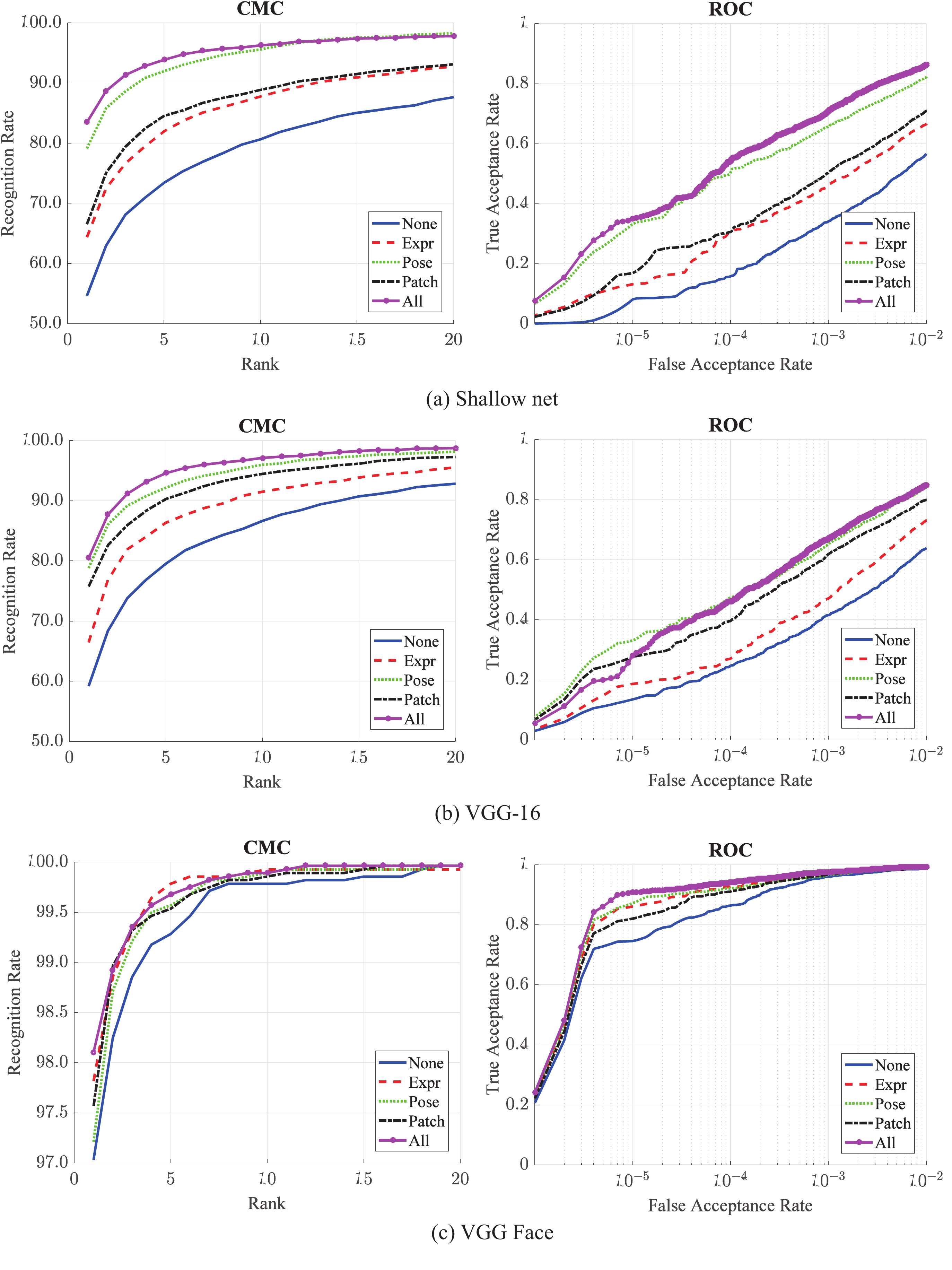}
		\caption{Evaluation of the augmentation methods. FRGC is used as a training set. The dataset was augmented using each augmentation method: expression, pose, patch, and all combined. The three CNNs ((a),(b), and (c)) were evaluated on the Bosphorus database.
		}
		\label{eval1}
		\vspace{-1em}
	\end{figure}
	
	\begin{figure}[t]
		\centering
		\includegraphics[width = 0.5\textwidth]{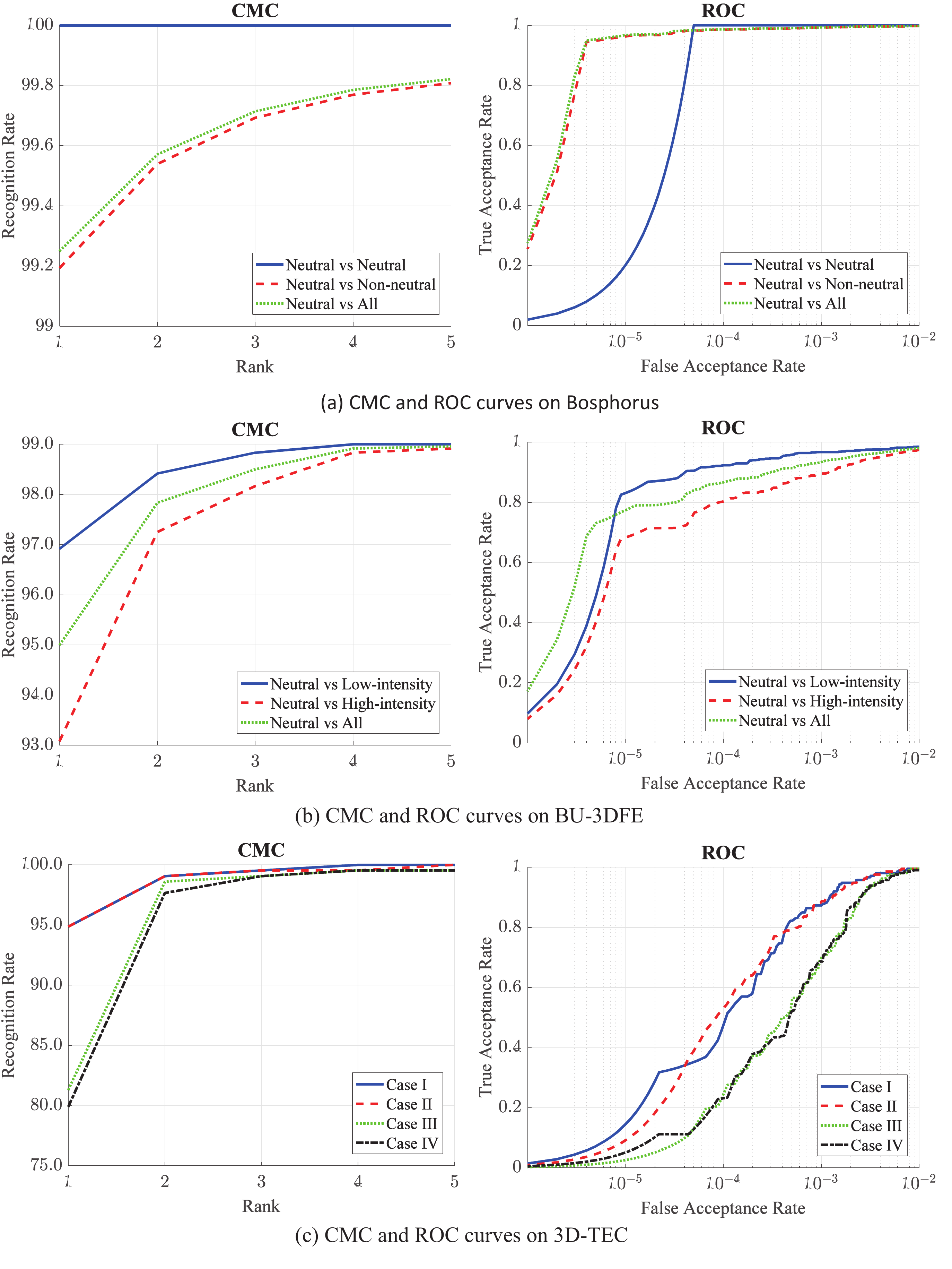}
		\caption{Evaluation results on the three databases. (a) and (b) 
			Performances were evaluated on three cases by varying expression intensities in a probe set. (c) Performances were evaluated on the four standard experiment cases.} 
		\vspace{-1em}
		\label{eval2}
	\end{figure}
	Figure \ref{eval1} shows the ROC and CMC curves from the experiments. Each augmentation method improves performances on the three CNNs. 
	Performances are largely increased in (a) and (b). Although there are small improvements in (c), it is more important because the performance without augmentation is already very high (rank-1 of 97.0\%). We also find that all CNNs gave the best rank-1 accuracy when a CNN is trained on the data using all of the three augmentation methods. In (a) and (b), pose augmentation shows the biggest improvements in the CMC curves. In (c), expression augmentation achieves the highest increase compared with other augmentation methods in the CMC curve.
	
	\subsection{Performances on the 3D Databases}
	We evaluate the proposed method on the Bosphorus~\cite{savran2008bosphorus}, BU-3DFE~\cite{yin20063d} and 3D-TEC~\cite{vijayan2011twins}. Each gallery set of the databases except 3D-TEC is augmented and used for training data. We follow evaluation protocols described in~\cite{li2014expression}. Figure \ref{eval2} shows ROC and CMC curves on the three databases and Table \ref{Rank-1} compares rank-1 accuracy with the state-of-the-art methods which have same protocols.
	
	{\bf Performances on Bosphorus}
	We set three experimental scenarios for detail analysis: (1) Neutral (105 galleries) vs Neutral (194 probes), (2) Neutral vs Non-neutral (2603 probes), and (3) Neutral vs All (2797 probes). The gallery sets are the same in the experiments. Our method achieves 100\%, 99.2\%, and 99.24\% rank-1 accuracies respectively. The rank-1 of 99.24\% in the experiment (3) is the highest accuracy compared with the state-of-the-art methods as shown in the Table \ref{Rank-1}. With the training set from only augmented FRGC dataset, we obtain rank-1 of 98.1\% as shown in Figure \ref{eval1}(c).
	
	{\bf Performances on BU-3DFE}
	We set three experimental scenarios depending on the intensity of expressions: (1) Neutral (100 gallery) vs Low-intensity (1200 probes), (2) Neutral vs High-intensity (1200 probes), and (3) Neutral vs All (2400 probes). The gallery sets are the same in the experiments. We obtain 97\%, 95\%, and 93\% rank-1 accuracies at each scenario. We have the second highest rank-1 accuracy in the Table \ref{Rank-1}. 
	
	{\bf Performances on 3D-TEC}
	We evaluate our method on 3D-TEC in the 4 cases described in~\cite{vijayan2011twins}. The rank-1 performances are quite low in Case \RNum{3},\RNum{4} compared with Case \RNum{1}, \RNum{2}. This is because the expression of a probe is different from one's face in the gallery but the same as the twin's expression in the gallery. Other methods except~\cite{li2014expression} show the same tendency as shown in Table \ref{Rank-1}. However, the rank-2 performances were largely increased by over 15\% in Case \RNum{3}, \RNum{4} and the rank-2 recognition rate is similar on all four cases. 
	\begin{table*}
		\begin{center}
			\begin{threeparttable}
				\begin{tabular}{|l|l|c|c|c|c|c|c|c|}
					\hline 
					\multirow{2}{*}{Approaches} & \multirow{2}{*}{Training data} & \multirow{2}{*}{Bosphorus} & \multirow{2}{*}{BU-3DFE} & \multicolumn{4}{l|}{3D-TEC }\\
					\cline{5-8}
					& & & & Case \RNum{1} & Case \RNum{2} & Case \RNum{3} & Case \RNum{4}\\
					\hline \hline
					Lei \etal~\cite{lei2016two} (2016) & Gallery\tnote{a} & 98.9 & 93.2 & - &- &- &-\\ \hline
					Ocegueda \etal~\cite{ocegueda2011ur3d} (2011) & FRGC v2 & 98.6 & {\bf 99.3} & - &- &- &- \\ \hline
					\multirow{2}{*}{Li \etal~\cite{li2014expression} (2014)} & BU-3DFE, Gallery\tnote{a} & 95.4 & - & 93.9 & 96.3 & \bf{90.7} & 91.6 \\ \cline{2-8} 
					& Bosphorus, Gallery\tnote{a}& - & 92.2 & 94.4 & {\bf96.7} & \bf{90.7} & \bf{92.5} \\ \hline
					Li \etal~\cite{li2015towards} (2015)& Gallery\tnote{a} & 98.8 & - & -& - & - & - \\
					\hline
					Berretti \etal~\cite{berretti2013matching} (2013) & BU-3DFE\tnote{b} & 95.7 & 87.5 & - & - & - & -\\
					
					\hline
					Huang \etal~\cite{huang2011novel} (2011) & None & - & - & 91.1 & 93.5 & 77.1 & 78.5 \\ \hline
					Huang \etal~\cite{huang2011textured} (2011) & FRGC v1 & - & - & 91.6 & 93.9 & 68.7 & 71.0 \\
					\hline
					Faltemier \etal~\cite{faltemier2008region} (2008) & FRGC v1 &- & - & 94.4 & 93.5 & 72.4 & 72.9 \\
					\hline
					\multirow{2}{*}{Ours} & FRGC\tnote{c}, Casia 3D\tnote{c} & \multirow{2}{*}{{\bf 99.2}} & \multirow{2}{*}{95.0} & \multirow{2}{*}{\bf{94.8}} & \multirow{2}{*}{94.8} & \multirow{2}{*}{81.3} & \multirow{2}{*}{79.9} \\ 
					& Gallery\tnote{a,d} & & & & & & \\
					\hline
				\end{tabular}
				\begin{tablenotes}
					\item[a] Corresponding gallery set of the testing set respectively.
					\item[b] A subset of BU-3DFE. 
					\item[c] Augmented entire data on expressions, poses, and random patches.
					\item[d] Augmented gallery sets of Bosphorus and BU-3DFE.
					\item[\_] Not reported
				\end{tablenotes}
			\end{threeparttable}
		\end{center}
		\caption{Comparision of rank-1 accuracy (\%) on public 3D face databases}
		\label{Rank-1}
		\vspace{-1em}
	\end{table*}
	\begin{table}
		\begin{center}
			\begin{threeparttable}
				\begin{tabular}{|l|c|c|c|c|}
					\hline
					Approaches & Processing & Matching & Total \\
					\hline\hline
					Spreeuwers ~\cite{spreeuwers2011fast} & 2.5 & 0.04 & {\bf 2.54} \\
					Lei \etal~\cite{lei2016two} & 6.08 & 2.41 & 8.49 \\
					Li \etal~\cite{li2014expression} & 3.05 & 0.5 & 3.55 \\
					Alyuz \etal~\cite{alyuz2010regional} & 36 & 0.02 & 36.02 \\
					Kakadiaris \etal~\cite{kakadiaris2007three} & 15 & 0.5 &  15.5 \\
					Li \etal~\cite{li2015towards}\tnote{a} & 69.5 & 5.5 &  75 \\
					Ours & 3.16 & 0.09 & 3.25 \\
					\hline
				\end{tabular}
				\begin{tablenotes}
					\item[a] Computation times when the gallery size is 105.
				\end{tablenotes}
			\end{threeparttable}
		\end{center}
		\caption{Comparision of computation time (s) of feature extraction, matching per a probe for identification.}
		\label{TimesComparision}
		\vspace{-1em}
	\end{table}
	
	From the results, our method achieves comparable performances to state-of-the-art methods on the three databases and can handle rich variations of expression. However, it is still hard to identify twins with different expressions.
	
	\subsection{Time Complexity Analysis}
	We evaluate our method on a PC with 2.6 GHz dual-processors and a NVIDIA K40 GPU for training and testing by using Caffe implementation~\cite{jia2014caffe}. In a 3D face recognition system, face identification is usually slow as a probe face needs to be matched with the whole gallery set. We measured the computation time for pre-processing and matching for identification per probe where a gallery size is 466. In this experiment, pre-processing includes time for processing the raw 3D data and extracting features. Table \ref{TimesComparision} shows computation time of our method and other methods. The time for our method takes $3.25$ seconds to identify a probe. Our time-consuming part is registration of a probe to a reference. It takes around $3$ seconds to register a probe using the rigid-ICP. The rest of the process (feature extraction and matching) can be done in less than a second. In a matching step, our method only needs to measure a cosine distance of N-dimensional feature vectors where $N$ is the size of a gallery set. Spreeuwers~\cite{spreeuwers2011fast} shows the computationally best method but the method is not carefully evaluated. This paper uses only one database for evaluation. Li \etal~\cite{li2014expression} introduced a method that takes $3.55$s to identify a probe when the size of the gallery is $466$. Since it involves an optimization ($l_0$ minimization) where computation times depends on the size of the gallery, it is hard to estimate its time complexity when the gallery size increases. Ocegueda \etal~\cite{ocegueda2011ur3d} did not report their time complexity. However, the pre-processing in \cite{ocegueda2011ur3d} takes at least 15s as it uses a deformable model fitting reported in ~\cite{kakadiaris2007three}.
	
	\section{Conclusions} \label{Conclusions}
	
	In this paper, we propose the first 3D face recognition model with a deep convolutional network (DCNN). 
	We show a prominent possibility of using a DCNN, even with a limited dataset. In order to overcome the challenge of acquiring large scale 3D facial datasets, we leverage fine-tuning and augmentation methods. Our method only requires standard pre-processing methods, including a nose tip detection and ICP, and does not involve complex feature extraction and matching, so that it is time efficient. 
	Our method is evaluated on the public 3D databases \cite{savran2008bosphorus, yin20063d, yin20063d} shows comparable performances to the state-of-the-art results in terms of accuracy while being more scalable. 
	
	{\small
		\bibliographystyle{ieee}
		\bibliography{egbib}
	}
	
\end{document}